\documentclass[journal]{IEEEtran}
\ifCLASSOPTIONcompsoc
  \usepackage[nocompress]{cite}
\else
  \usepackage{cite}
\fi

\ifCLASSINFOpdf
  \usepackage[pdftex]{graphicx}
\else
\fi

\usepackage{url}

\usepackage{algorithm}
\usepackage{algpseudocode}

\usepackage{xcolor}

\begin{document}
%
\title{Combining Optimal Path Search With Task-Dependent Learning in a Neural Network}
%
%
\author{Tomas Kulvicius,
        Minija Tamosiunaite,
        and Florentin W\"org\"otter
\thanks{The research leading to these results has received funding from the  European Community's H2020 Programme (Future and Emerging Technologies, FET) under grant agreement no. 899265 (ADOPD), and the German Science Foundation (DFG), under grant agreement no. WO 388/16-1.}
\thanks{T. Kulvicius, M. Tamosiunaite and F. W\"org\"otter are with the Department for Computational Neuroscience, University of G\"ottingen, 37073 G\"ottingen, Germany.}
\thanks{T. Kulvicius is also with University Medical Center G\"ottingen, Systemic Ethology and Developmental Science, Child and Adolescent Psychiatry and Psychotherapy, G\"ottingen, Germany.}
\thanks{M. Tamosiunaite is also with the Faculty of Computer Science, Vytautas Magnus University, 53361 Akademija, Kauno r., Lithuania.}
}

\maketitle

\begin{abstract}
Finding optimal paths in connected graphs requires determining the smallest total cost for traveling along the graph's edges. This problem can be solved by several classical algorithms where, usually, costs are predefined for all edges. Conventional planning methods can, thus, normally not be used when wanting to change costs in an adaptive way following the requirements of some task. Here we show that one can define a neural network representation of path finding problems by transforming cost values into synaptic weights, which allows for online weight adaptation using network learning mechanisms. When starting with an initial activity value of one, activity propagation in this network will lead to solutions, which are identical to those found by the Bellman-Ford algorithm. The neural network  has the same algorithmic complexity as Bellman-Ford and, in addition,  we can show that network learning mechanisms (such as Hebbian learning) can adapt the weights in the network augmenting the resulting paths according to some task at hand. We demonstrate this by learning to navigate in an environment with obstacles as well as by learning to follow certain sequences of path nodes. Hence, the here-presented novel algorithm may open up a different regime of applications where path-augmentation (by learning) is directly coupled with path finding in a natural way.
\end{abstract}


\section{Introduction}

This study addresses the so-called single-source shortest path (SSSP) problem. Possibly the most prominent goal for any path finding problem is to determine a path between two vertices (usually called source and destination) in a graph such that the total summed costs for traveling along this path is minimised. Usually (numerical) costs are associated to the edges that connect the vertices and costs are this way accumulated when traveling along a set of them. In the SSSP problem, the aim is to find the shortest paths from one vertex (source) to all other remaining vertices in the graph.

The SSSP problem has a wide range of applications, e.g., in computer networks (shortest path between two computers), social networks (shortest path between two persons or linking between two persons), trading and finance (e.g., currency exchange), multi-agent systems such as games, task and path planning in robotics, etc., just to name a few.

The most general way to solve the SSSP problem is the Bellman-Ford (BF) algorithm \cite{Bellman1958,Ford1956,Baras2010}, which can also deal with graphs that have some negative cost values. In this work, we present a neural implementation which is mathematically equivalent to the BF algorithm with an algorithmic complexity which is the same as for BF.

The neural implementation relies on the multiplication of activities (instead of adding of costs). As a consequence, it is directly compatible with (Hebbian) network learning, which is not the case for any additively operating path planning algorithm. To demonstrate this, we are using Hebbian type 3-factor learning \cite{Hebb1949,Kusmierz2017,Porr2007} to address SSSP tasks under some additional constraints, solved by the 3-factor learning.

Our paper is structured as follows. First, we will provide an overview of the state of the art methods and state our contribution with respect to that. Next, we will describe details of the BF algorithm and the proposed neural network (NN-BF). This will be followed by a comparison between the BF algorithm and NN-BF and by examples of combining NN-BF based planning with three-factor learning. Finally, we will conclude our study with a summary and provide an outlook for future work.

\section{Related Work}

\subsection{State-of-the-art}

There are mainly two types of approaches to solve the SSSP problem: classical algorithms and approaches based on artificial neural networks. Classical algorithms exist with different complexity and properties. The simplest and fastest algorithm is breadth-first search (BFS) \cite{Moore1959,Merrill2012}. However, it can only solve the SSSP problem for graphs with uniform costs, i.e., all costs equal to 1. Dijkstra's algorithm \cite{Dijkstra1959} as well as the Bellman-Ford (BF) algorithm \cite{Bellman1958,Ford1956} can deal with graphs that have arbitrary costs. From an algorithmic point of view, Dijkstra's algorithm is faster than the BF algorithm, however, Dijkstra only works on graphs with positive costs, whereas the BF algorithm can also solve graphs where some of the costs are negative. Furthermore, Dijkstra is a greedy algorithm and requires a priority queue, which makes it not so well suited for parallel implementation as compared to BF.

It is worth noting, that there are two versions of the BF algorithm (see \cite{Baras2010}). Both versions have the same basic algorithmic complexity but version 2 operates on graph nodes and not on graph edges (as version 1), which allows implementing version 2 in a totally asynchronous way where computations at each node are independent of other nodes \cite{Baras2010}. In spite of this, version 2 is largely ignored in the literature and users will usually be directed to version 1 when doing a (web-)search.

Some heuristic search algorithms such as A* \cite{Hart1968} and its variants (see \cite{Korf1985,Koenig2004,Sun2008,Harabor2011}) exist, which are faster than Dijkstra or BF, but they can only solve the single-source single-target shortest path problem.

The most general, but slowest, algorithm to solve the SSSP problem is the Floyd-Warshall algorithm \cite{Floyd1962,Warshall1962}, which finds all-pairs shortest paths (APSP), i.e., all shortest paths from each vertex to all other vertices. Another way to solve the APSP problem is by using Johnson's algorithm \cite{Johnson1977}, which utilises Bellman-Ford and Dijkstra's algorithms and is -- under some conditions -- faster than the Floyd-Warshall algorithm (essentially this is the case for sparse graphs).

Many different algorithms exist, which utilise artificial neural networks to solve shortest path problems. Some early approaches, which were dealing with relatively small graphs (below 100 nodes), were based on Hopfield networks \cite{Ali1993,Park1998,Araujo2001} or Potts neurons and a mean field approach \cite{Hakkinen1998}. These approaches, however, may not always give optimal solutions or may fail to find solutions at all, especially for larger graphs.

Some other bio-inspired neural networks were proposed \cite{Glasius1995,Glasius1996,Bin2004,Yang2001,Qu2009,Ni2017,Kulvicius2021a} for solving path planning problems. These approaches work on grid structures, where activity in the network is propagated from the source neuron to the neighbouring neurons until activity propagation within the whole network is finished. Shortest paths then can be reconstructed by following the activity gradients. The drawback of these approaches is, however, that they are specifically designed for grid structures and can not be applied at general graphs with arbitrary costs.

Several deep learning approaches were proposed to solve shortest path problems, for example using a deep multi-layer perceptron (DMLP, \cite{Qureshi2018}), fully convolutional networks \cite{Perez2018,Ariki2019,Kulvicius2021}, or a long short-term memory (LSTM) network \cite{Bency2019}. Inspired by classical reinforcement learning algorithms (RL), e.g., Q-learning, SARSA \cite{Sutton2018}, some deep reinforcement learning algorithms had been employed \cite{Tai2017,Panov2018,Banino2018} too. These approaches are designed to solve path planning problems in 2D or 3D spaces and cannot deal with graphs with arbitrary costs.

In addition to this, deep learning approaches based on graph neural networks (GNNs) have been employed to solve path problems, too \cite{Schlichtkrull2018,Zhang2018,Velickovic2020,Teru2020,Zhu2021,Yang2021}, mostly in the context of relation and link predictions. While deep learning approaches may lead to a better run-time performance as compared to classical approaches due to fast inference, all deep learning based approaches need to learn their (many) synaptic weights before they are functional. This usually requires a large amount of training data. Another disadvantage of these approaches is that optimal solution is not guaranteed since networks intrinsically perform function approximation based on the training data. Moreover, in some of the cases networks may even fail to find a solution at all, especially, in cases where training data is quite different from new cases (generalisation issue for out-of-distribution data).

Recently graph-based neural networks have been proposed too, such as spatial-temporal graph convolutional networks (ST-GCN) for skeleton based action recognition\cite{Yan2018},  spatial-temporal graph network for video supported machine translation\cite{Li2023}, graph transformer network (TN-ZSTAD) for zero-shot temporal activity detection from videos\cite{Zhang2023}. Other deep networks are based on scene graph generation (SGG), e.g., CRF-based SGG\cite{Cong2018}, TransE-based SGG \cite{Bordes2013}, CNN-based SGG\cite{Woo2018}, RNN/LSTM-based SGG\cite{Xu2017,Zellers2018,Gao2019}, and GNN-based SGG\cite{Li2018,Yang2018}. While these approaches have been mostly applied for text and image/video analysis, such as text-to-image generation, image/video captioning, image retrieval, scene understanding, and human-object interaction (for a comprehensive survey see \cite{Chang2023}), none of them have been exploited to specifically address SP/SSSP problems.

Furthermore, some approaches exist for an automated search (tuning) of generative adversarial network (GAN) architectures such as neural architecture search (NAS)\cite{Zhang2020,Li2022}, AutoGAN\cite{Gong2019}, AGAN\cite{Wang2019} and for the simultaneous learning of network architecture parameters and weights (ZeroNAS, \cite{Yan2022}). Although these approaches are related to our proposed method, there are two principle differences. First, the afore mentioned approaches are used to optimise network architectures or architecture and weights to perform discriminating tasks, e.g., image classification or object detection, whereas in our approach we are dealing with the SSSP problem, where we build the network architecture and update connection weights based on local learning rules. Hence, we do not specifically optimise network architecture and weights to map input to outputs. Second, in contrast to NAS approaches, weights in our network correspond to costs, which allow using activity propagation within the network to find optimal solutions for the SSSP tasks.

\begin{figure}[ht!]
\centering \includegraphics[width=0.98\linewidth]{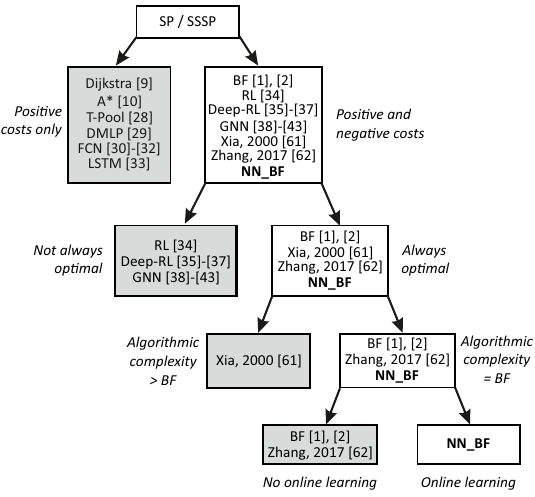} \caption{Comparison of different methods for solving shortest-path (SP) and single-source shortest paths (SSSP) problems.}
\label{method_tree}
\end{figure}

\begin{table*}[ht!]
\caption{Comparison of different methods for solving shortest-path (SP) and single-source shortest paths (SSSP) problems.}
\begin{scriptsize}
\begin{tabular}{|lllllllll|}
\hline
\multicolumn{1}{|l|}{\textbf{Algorithm}}                                                                     & \multicolumn{1}{l|}{\begin{tabular}[c]{@{}l@{}}\textbf{Problem:}\\ \textbf{SP/SSSP}\end{tabular}} & \multicolumn{1}{l|}{\begin{tabular}[c]{@{}l@{}}\textbf{Numerical}\\\textbf{calculation basis}\end{tabular}}             & \multicolumn{1}{l|}{\textbf{Cost type}}                                                         & \multicolumn{1}{l|}{\begin{tabular}[c]{@{}l@{}}\textbf{Algorithmic}\\ \textbf{procedure}\end{tabular}}                                                                  & \multicolumn{1}{l|}{\begin{tabular}[c]{@{}l@{}}\textbf{Training} \end{tabular}} & \multicolumn{1}{l|}{Optimality}                                                          & \multicolumn{1}{l|}{\textbf{Complexity}}                                                                & \begin{tabular}[c]{@{}l@{}}\textbf{Online learning}\end{tabular}               \\ \hline
\multicolumn{1}{|l|}{A*\cite{Hart1968}}                                                                             & \multicolumn{1}{l|}{SP}                                                         & \multicolumn{1}{l|}{Cost values}                                                                       & \multicolumn{1}{l|}{Only positive}                                                     & \multicolumn{1}{l|}{\begin{tabular}[c]{@{}l@{}}Additive\\ cost summation\end{tabular}}                                                                & \multicolumn{1}{l|}{NO}                                                                  & \multicolumn{1}{l|}{Optimal}                                                             & \multicolumn{1}{l|}{$\mathcal{O}(bd)$}                                                                     & Not directly possible                                                              \\ \hline
\multicolumn{1}{|l|}{Dijkstra\cite{Dijkstra1959}}                                                                       & \multicolumn{1}{l|}{SSSP}                                                       & \multicolumn{1}{l|}{Cost values}                                                                       & \multicolumn{1}{l|}{Only positive}                                                     & \multicolumn{1}{l|}{\begin{tabular}[c]{@{}l@{}}Additive\\ cost summation\end{tabular}}                                                                & \multicolumn{1}{l|}{NO}                                                                  & \multicolumn{1}{l|}{Optimal}                                                             & \multicolumn{1}{l|}{\begin{tabular}[c]{@{}l@{}}$\mathcal{O}(|E|+|V|^2$) \\ or\\ $\mathcal{O}(|E|+$ \\ $|V|log|V|)$\end{tabular}} & Not directly possible                                                              \\ \hline
\multicolumn{1}{|l|}{BF\cite{Bellman1958,Ford1956}}                                                                             & \multicolumn{1}{l|}{SSSP}                                                       & \multicolumn{1}{l|}{Cost values}                                                                       & \multicolumn{1}{l|}{\begin{tabular}[c]{@{}l@{}}Positive and\\ negative\end{tabular}}   & \multicolumn{1}{l|}{\begin{tabular}[c]{@{}l@{}}Additive\\ cost summation\end{tabular}}                                                                & \multicolumn{1}{l|}{NO}                                                                  & \multicolumn{1}{l|}{Optimal}                                                             & \multicolumn{1}{l|}{$\mathcal{O}(|V||E|)$}                                                                 & Not directly possible                                                              \\ \hline
\multicolumn{1}{|l|}{\begin{tabular}[c]{@{}l@{}}Recurrent\\ neural\\ network \\Xia, 2000 \cite{Xia2000}\end{tabular}} & \multicolumn{1}{l|}{SP}                                                         & \multicolumn{1}{l|}{\begin{tabular}[c]{@{}l@{}}Weight values =\\ cost values\end{tabular}}             & \multicolumn{1}{l|}{\begin{tabular}[c]{@{}l@{}}Positive and\\ negative\end{tabular}}   & \multicolumn{1}{l|}{\begin{tabular}[c]{@{}l@{}}Additive/Multiplicative\\ activity propagation\end{tabular}} & \multicolumn{1}{l|}{NO}                                                                  & \multicolumn{1}{l|}{Optimal}                                                             & \multicolumn{1}{l|}{\begin{tabular}[c]{@{}l@{}}$\mathcal{O}(|V|^2)$\\ per interation\end{tabular}}          & Not shown                                                                          \\ \hline
\multicolumn{1}{|l|}{\begin{tabular}[c]{@{}l@{}}Neural\\ network \\ Zhang, 2017\cite{Zhang2017}\end{tabular}}           & \multicolumn{1}{l|}{SSSP}                                                       & \multicolumn{1}{l|}{\begin{tabular}[c]{@{}l@{}}Weight values =\\ cost values\end{tabular}}             & \multicolumn{1}{l|}{\begin{tabular}[c]{@{}l@{}}Positive and\\ negative\end{tabular}}   & \multicolumn{1}{l|}{\begin{tabular}[c]{@{}l@{}}Additive activity\\ propagation\end{tabular}}                & \multicolumn{1}{l|}{NO}                                                                  & \multicolumn{1}{l|}{Optimal}                                                             & \multicolumn{1}{l|}{Not stated}                                                                & Not shown                                                                          \\ \hline
\multicolumn{1}{|l|}{DMLP\cite{Qureshi2018}}                                                                           & \multicolumn{1}{l|}{SP}                                                         & \multicolumn{1}{l|}{Weight values*}                                                                    & \multicolumn{1}{l|}{n/a}                                                               & \multicolumn{1}{l|}{\begin{tabular}[c]{@{}l@{}}Additive/Multiplicative\\ activity propagation\end{tabular}} & \multicolumn{1}{l|}{YES}                                                                 & \multicolumn{1}{l|}{\begin{tabular}[c]{@{}l@{}}Optimality not\\ guaranteed\end{tabular}} & \multicolumn{1}{l|}{Not stated}                                                                & n/a                                                                                \\ \hline
\multicolumn{1}{|l|}{FCN\cite{Perez2018,Ariki2019,Kulvicius2021}}                                                                            & \multicolumn{1}{l|}{SP}                                                         & \multicolumn{1}{l|}{Weight values*}                                                                    & \multicolumn{1}{l|}{n/a}                                                               & \multicolumn{1}{l|}{\begin{tabular}[c]{@{}l@{}}Additive/Multiplicative\\ activity propagation\end{tabular}} & \multicolumn{1}{l|}{YES}                                                                 & \multicolumn{1}{l|}{\begin{tabular}[c]{@{}l@{}}Optimality not\\ guaranteed\end{tabular}} & \multicolumn{1}{l|}{Not stated}                                                                & n/a                                                                                \\ \hline
\multicolumn{1}{|l|}{LSTM\cite{Bency2019}}                                                                           & \multicolumn{1}{l|}{SP}                                                         & \multicolumn{1}{l|}{Weight values*}                                                                    & \multicolumn{1}{l|}{n/a}                                                               & \multicolumn{1}{l|}{\begin{tabular}[c]{@{}l@{}}Additive/Multiplicative\\ activity propagation\end{tabular}} & \multicolumn{1}{l|}{YES}                                                                 & \multicolumn{1}{l|}{\begin{tabular}[c]{@{}l@{}}Optimality not\\ guaranteed\end{tabular}} & \multicolumn{1}{l|}{Not stated}                                                                & n/a                                                                                \\ \hline

\multicolumn{1}{|l|}{RL\cite{Sutton2018}}                                                                        & \multicolumn{1}{l|}{SSSP}                                                     & \multicolumn{1}{l|}{Value tables}                                                                    & \multicolumn{1}{l|}{\begin{tabular}[c]{@{}l@{}}Rewards or\\ punishments\end{tabular}}                                                               & \multicolumn{1}{l|}{\begin{tabular}[c]{@{}l@{}}Reward\\ propagation\end{tabular}} & \multicolumn{1}{l|}{NO}                                                                 & \multicolumn{1}{l|}{\begin{tabular}[c]{@{}l@{}}Optimality not\\ guaranteed**\end{tabular}} & \multicolumn{1}{l|}{\begin{tabular}[c]{@{}l@{}}$\mathcal{O}(n^2)$ or\\ $\mathcal{O}(n^3)$\cite{Koenig1993}\end{tabular}}                                                                & Reinforcement learning                                                             \\ \hline

\multicolumn{1}{|l|}{Deep-RL\cite{Tai2017,Panov2018,Banino2018}}                                                                        & \multicolumn{1}{l|}{SSSP}                                                     & \multicolumn{1}{l|}{Weight values*}                                                                    & \multicolumn{1}{l|}{n/a}                                                               & \multicolumn{1}{l|}{\begin{tabular}[c]{@{}l@{}}Additive/Multiplicative\\ activity propagation\end{tabular}} & \multicolumn{1}{l|}{YES}                                                                 & \multicolumn{1}{l|}{\begin{tabular}[c]{@{}l@{}}Optimality not\\ guaranteed\end{tabular}} & \multicolumn{1}{l|}{Not stated}                                                                & Reinforcement learning                                                             \\ \hline

\multicolumn{1}{|l|}{GNN\cite{Schlichtkrull2018,Zhang2018,Velickovic2020,Teru2020,Zhu2021,Yang2021}}                                                                            & \multicolumn{1}{l|}{SSSP}                                                     & \multicolumn{1}{l|}{Weight values*}                                                                    & \multicolumn{1}{l|}{n/a}                                                               & \multicolumn{1}{l|}{\begin{tabular}[c]{@{}l@{}}Additive/Multiplicative\\ activity propagation\end{tabular}} & \multicolumn{1}{l|}{YES}                                                                 & \multicolumn{1}{l|}{\begin{tabular}[c]{@{}l@{}}Optimality not\\ guaranteed\end{tabular}} & \multicolumn{1}{l|}{Not stated}                                                                & n/a                                                                                  \\ \hline
\multicolumn{1}{|l|}{T-POOL\cite{Kulvicius2021a}}                                                                         & \multicolumn{1}{l|}{SSSP}                                                       & \multicolumn{1}{l|}{\begin{tabular}[c]{@{}l@{}}Weight values =\\ cost values\end{tabular}}             & \multicolumn{1}{l|}{\begin{tabular}[c]{@{}l@{}}Only positive\\ (uniform)\end{tabular}} & \multicolumn{1}{l|}{\begin{tabular}[c]{@{}l@{}}Additive/Multiplicative\\ activity propagation\end{tabular}} & \multicolumn{1}{l|}{NO}                                                                  & \multicolumn{1}{l|}{Optimal}                                                             & \multicolumn{1}{l|}{$\mathcal{O}(|V||L|)$}                                                                 & n/a                                                                                \\ \hline
\multicolumn{1}{|l|}{\begin{tabular}[c]{@{}l@{}}\textbf{NN-BF}\\\textbf{(this study)}\end{tabular}}                                                             & \multicolumn{1}{l|}{SSSP}                                                       & \multicolumn{1}{l|}{\begin{tabular}[c]{@{}l@{}}Weight values = \\ inverse of \\ cost values\end{tabular}} & \multicolumn{1}{l|}{\begin{tabular}[c]{@{}l@{}}Positive and\\ negative\end{tabular}}   & \multicolumn{1}{l|}{\begin{tabular}[c]{@{}l@{}}Multiplicative activity\\ propagation\end{tabular}}          & \multicolumn{1}{l|}{NO}                                                                  & \multicolumn{1}{l|}{Optimal}                                                             & \multicolumn{1}{l|}{$\mathcal{O}(|V||E|)$}                                                                 & \begin{tabular}[c]{@{}l@{}}Hebbian learning,\\ Reinforcement learning\end{tabular} \\ \hline
\multicolumn{9}{|l|}{\begin{tabular}[c]{@{}l@{}}SP -- single-source -- single-target;  SSSP – Sigle-Source Shortest Paths (single-source -- multi-targets)\\
$b$ -- branching factor; $d$ -- depth of the solution: $|V|$ -- number of nodes, $|E|$ -- number of edges, $|L|$ -- number of layers\\
$^*$Weight values do not correspond to the costs \\ $^{**}$RL usually does not converge to optimal solutions in high dimensional spaces within reasonable time\end{tabular} } \\ \hline
\end{tabular}
\end{scriptsize}
\label{method_comp}
\end{table*}

\subsection{Contribution}

The comparison of the above discussed approaches for solving SP/SSSP problems is graphically summarized in Fig.~\ref{method_tree} and details are provided in Table~\ref{method_comp}. Here we compare our approach with respect to the following criteria: 1) ability to deal with positive and negative costs, 2) solution optimality, and 3) algorithmic complexity. Given these criteria, one can see that our approach most closely relates to the neural approaches by Xia\&Wang\cite{Xia2000}, and Zhang\&Li\cite{Zhang2017}. The approach by Xia\&Wang\cite{Xia2000} utilises a relatively complex recurrent neural network architecture. It can can deal with optimal and negative costs and find optimal solutions but it has a higher complexity as compared to the Bellman-Ford (BF) algorithm. Zhang\&Li\cite{Zhang2017}, as in our approach, proposed a neural implementation of the BF algorithm with the same complexity, however, it uses additive activity propagation (akin to additive cost propagation in the BF algorithm) as compared to multiplicative activity propagation as proposed by our algorithm. Moreover, neither Xia\&Wang\cite{Xia2000} nor Zhang\&Li\cite{Zhang2017} show a combination of addressing the shortest path problem together with neural learning.

In summary, we present a neural implementation of the Bellman-Ford algorithm, which finds optimal solutions for graphs with arbitrary positive and (some) negative costs with the same algorithmic complexity as BF. The advantage of this is that, different from BF and other neural implementations of the BF algorithm \cite{Xia2000,Zhang2017}, we show that we can also directly apply neuronal learning rules that can dynamically change the activity in the network and lead to re-routing of paths according to some requirements. We show this by two cases where neuronal learning is used to structure the routing of the paths in a graph in different ways.

\section{Methods}

\subsection{Definitions}

We denote a weighted graph as $G(V, E, C)$ with vertices $V$, edges $E$ and corresponding edge costs $C$. In the first version of the BF algorithm a so called relaxation procedure (cost minimisation) is performed for all edges whereas in the second version of the BF algorithm relaxation is performed for all nodes. The relaxation procedure is performed for a number of iterations until convergence, i.e., approximate distances are replaced with shorter distances until all distances finally converge.

The BF algorithm is guaranteed to converge if relaxation of all edges is performed $|V|-1$ times\footnote{Note that this corresponds to the worst case. See also algorithmic complexity analysis below.}. After convergence, shortest paths (sequences of nodes) can be found from a list of predecessor nodes backwards from target nodes to the source node.

\subsection{Neural implementation of the Bellman-Ford algorithm}

In this section we will describe our neural implementation of Bellman-Ford (BF) algorithm. Similar to BF, the proposed neural network finds shortest paths in a weighted graph from a source node (vertex) to all other nodes, however, shortest paths here are defined in terms of maximal activations from the source neuron to all other neurons.

\begin{algorithm*}
\caption{Pseudo-code of neural network algorithm.}
\label{alg_nn}
\textbf{Input:} list of neuron connections $(i, j)$ with weights $(w_{i,j})$, source neuron $(s)$, and costs $(c_{i,j})$ \\
\textbf{Output:} list of neuron activations (outputs), list of input nodes with maximal activity \\
\begin{algorithmic}
\For {each neuron $i$}
    \State $w[i,j] \gets 1 - c[i,j]/K$ \Comment{Initialise connection weights based on costs}
    \State $activations[i] \gets 0$ \Comment{Initialise all neuron activations with $0$}
    \State $max\_inputs[i] \gets null$ \Comment{Initialise all maximal input nodes with $null$}
\EndFor
\State $activations[s] \gets 1$ \Comment{Initialise activation of the source neuron with $1$}
\\
\For {$k \gets 1$ to $N-1$} \Comment{Repeat $N-1$ times ($N$ - number of neurons)}
    \For {each neuron $i$} \Comment{For each neuron compute its activation}
        \For {each input $j$ of neuron $i$ with weight $w[i,j]$}
            \State $weighted\_input \gets activations[j] \times w[i,j]$
                \If {$weighted\_input > activations[i]$} 
                    \State $activations[i] \gets weighted\_input$
                    \State $max\_inputs[i] \gets j$
                \EndIf
        \EndFor
    \EndFor
    \State $activations[s] \gets 1$ \Comment{Set activation of the source neuron to $1$}
\EndFor \\
\\
\Return{$activations$, $max\_inputs$} \Comment{Return list of activations and list of max inputs}
\end{algorithmic}
\end{algorithm*}

The neural network has $N$ neurons where $N$ corresponds to the number of vertices $|V|$ in a graph, connections between neurons $i$ and $j$ ($i,j = 1 \dots N$) correspond to the edges, and connection weights correspond to the (inverse) costs of edges which are computed by
\begin{equation}
 w_{i,j} = 1 - \frac{c_{i,j}}{K},
\end{equation}
where $K>0$. We can prove that the solutions obtained with the neural network are identical to the solutions obtained with BF under certain constraints (see section~\ref{sec_solution_equality}).

\begin{figure*}[ht!]
\centering \includegraphics[width=0.95\linewidth]{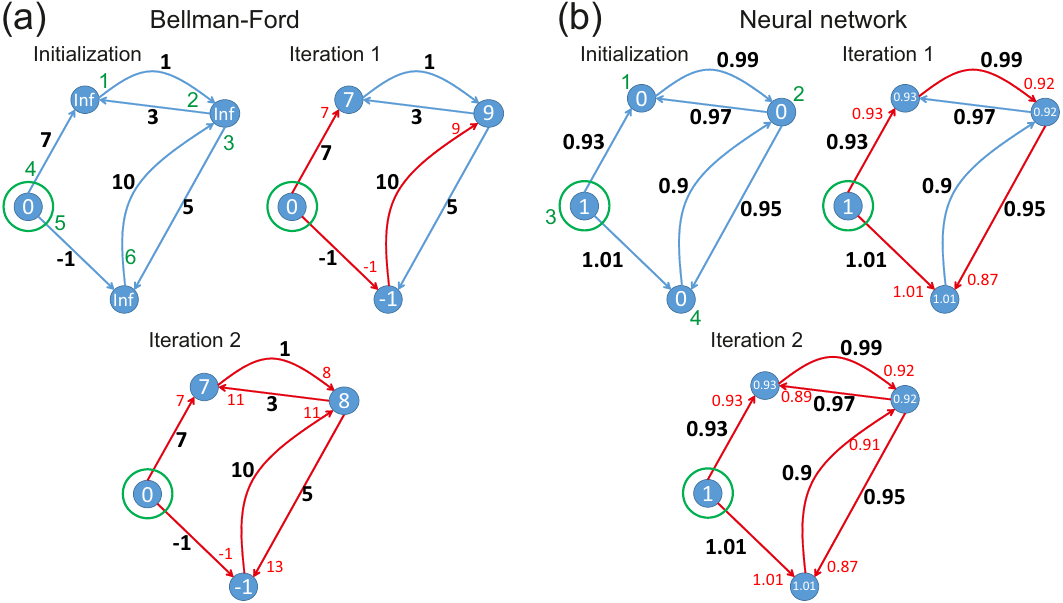}
\caption{Example of solving a directed weighted graph with four vertices and six edges using \textbf{(a)} BF and \textbf{(b)} NN-BF. Green circle denotes the source vertex/neuron. Black numbers correspond to the edge/connection weights. Red numbers correspond to costs/activations from the source vertex/neuron for the respective edges/inputs. Numbers inside graph vertices/neurons correspond to costs/activations from the source to the respective nodes/neurons. We used $K=100$ to convert edge costs to connections weights, i.e., $w_i = 1 - c_i/100$ ($i=1\dots6$). Green numbers in (a) correspond to the numbers of edges and the order in which they were processed, whereas green numbers in (b) correspond to numbers of neurons and their order in which their were processed.}
\label{bf_net}
\end{figure*}

We present the pseudo-code of the neural network algorithm (NN-BF)  in Algorithm~\ref{alg_nn}. Similar to BF version 2 (see \cite{Baras2010}), we run NN-BF for a number of iterations until convergence operating on the graph nodes (neurons). In every iteration we update the activation (output) of each neuron by
\begin{equation}
\label{eq_out}
 a_i = \max_j \{a_j \, w_{i,j}\},
\end{equation}
where $a_j$ correspond to the inputs of neuron $i$, and $w_{i,j}$ to the connection weights between input neurons $j$ and output neuron $i$. To start this process, we set the activation of the source neuron to $1$. Thus, this way activity from the source neuron is propagated to all other neurons until activation has converged everywhere. After convergence, shortest paths (sequences of neurons) can be found from a list of maximal input activations backwards from target neurons to the source neuron (similar to BF algorithm).

The NN-BF algorithm is guaranteed to converge if activity of the network is updated $N-1$ ($N = |V|$) times. This can be shown by the the worst case scenario where neurons would be connected in a sequence, i.e., $1 \to 2\to 3 \to \dots \to N$. Thus, to propagate activity from the $1st$ neuron to the last neuron in the sequence one would need to perform $N-1$ updates. 

A comparison of solving a simple graph using BF (version 1) and NN-BF is presented in Fig.~\ref{bf_net}. Here we used a simple directed weighted graph with four nodes and six edges with one negative cost and five positive costs. We show the resulting distances in (a) and network activations in (b) after each iteration (numbers inside the nodes/neurons of the graph/neural network). Red edges/connections denote distance/activity propagation from the source node/neuron to all other nodes/neurons. For more details please refer to the figure caption. In this case, the algorithm converged after two iterations when using BF  and after one iteration when using NN-BF.

\subsection{Performance evaluation of the neural planner}

We evaluated the performance of the neural network and compared it against  performance of the Bellman-Ford algorithm (version 1) with respect to 1) solution equality, 2) algorithmic complexity and 3) number of iterations until convergence.

For the numerical evaluation we generated random directed weighted graphs of different density, from sparse graphs (only few connections per node) to fully connected graphs (every node connects to all other nodes). To generate graphs of different density, we used the probability $p_e$ to define whether two nodes will be connected or not. For example, a graph with 100 nodes and $p_e = 0.1$ will obtain a $10\%$ connectivity rate, and, thus, to $\approx 10$ edges per node and $\approx 1,000$ edges in total. 

Costs for edges were assigned randomly from specific ranges (e.g., $[-10,-1]$ and $[1,100]$), where negative weights were assigned with probability $p_n$ (e.g., $p_n = 0.1$) and positive weights with probability $1-p_n$. The specific parameters for each evaluation case will be provided in the results sections.

\subsection{Combining neural planning network with neural plasticity}

The neural implementation of the Bellman-Ford algorithm allows us to use neuronal learning rules in order to dynamically change connection weights between neurons in the network and, thus, the outcome of the neural planner. The methods applied to achieve this will be presented directly in front of the corresponding results to allow for easier reading.

\section{Results}
\subsection{Comparison of BF and NN-BF}
In the following we first provide a comparison between BF and NN-BF, and then we will show results for NN-BF also on network learning.

\subsubsection{Solution equality \label{sec_solution_equality}}

To prove equivalence one can pairwise compare two paths A and B between some source and a target, where A is the best path under the Bellman-Ford condition and B is any another path. Path A is given by nodes $a_i$ and B by $b_i$. As described above, for a Bellman-Ford cost of $a_i$ we define the weight of the NN-BF connection as $w_i = 1-\frac{a_i}{K}$ and likewise for $b_i$. We need to show that:
\begin{equation}\label{ind_D1}
\textrm{if}~\sum_{i=1}^{n} a_i <  \sum_{i=1}^{n+m} b_i ~\textrm{then}~
\prod_{i=1}^{n}(1-\frac{a_i}{K}) > \prod_{i=1}^{n+m}(1-\frac{b_i}{K})
\end{equation}

Note, in general we can assume for both sides a path length of $n$, because -- if one path is shorter than the other -- we can just extend this by some dummy nodes with zero cost such that the Bellman-Ford costs still add up correctly to the total.

\noindent
First we analyse the simple case of $n=2$. The general proof, below, makes use of the structural similarity to the simple case. Thus, we show:
\begin{eqnarray}
\nonumber  a_1 + a_2       &<&  b_1 + b_2 \Leftrightarrow\\
 (1-\frac{a_1}{K}) (1-\frac{a_2}{K}) &>& (1-\frac{b_1}{K}) (1-\frac{b_2}{K}),
\label{start_condition}
\end{eqnarray}
\noindent
We start from the second line in Eq.~\ref{start_condition} and rewrite it as:
\begin{equation}
{\frac{1}{K^2}}[K^2 - K(a_1 + a_2) + a_1 a_2] > {\frac{1}{K^2}}[K^2 - K(b_1 + b_2) + b_1 b_2]
\label{l_out}
\end{equation}
\noindent
We simplify and divide by $K$, where $K>0$, to:
\begin{equation}
-a_1 - a_2+{\frac{a_1 a_2}{K}} > -b_1 - b_2 +{\frac{b_1 b_2}{K}}
\label{simp_case}
\end{equation}
Now we let $K\rightarrow \infty$ and get:
\begin{equation}
a_1 + a_2 < b_1 + b_2
\end{equation}
\noindent
which proves conjecture Eq.~\ref{start_condition} for large enough $K$.

\noindent
To generalize this we need to show that Eq.~\ref{ind_D1} holds. For  this, we analyse the second inequality given by:
\begin{equation}\label{second_ineq}
\prod_{i=1}^{n}(1-\frac{a_i}{K}) > \prod_{i=1}^{n}(1-\frac{b_i}{K})
\end{equation}

\noindent
From the structure of Eq.~\ref{l_out} we can deduce that:
\begin{eqnarray}\label{prod_expd}
\prod_{i=1}^{n}(1-\frac{a_i}{K}) = {\frac{1}{K^n}}(K^n - K^{n-1}\sum_i a_i + \nonumber \\ K^{n-2} \sum_{i\ne j} a_i a_j - K^{n-3} \sum_{i\ne j \ne k} a_i a_j a_k \pm \dots )
\end{eqnarray}
\noindent
and likewise for $b_i$, which allows writing out both sides of inequality Eq.~\ref{ind_D1}. When doing this (not shown to save space), we see that the fore-factor $\frac{1}{K^n}$ can be eliminated. Then the term $K^n$ also subtracts away from both sides. After this we can divide the remaining inequality as in Eq.~\ref{simp_case} above, here using $K^{n-1}$ as divisor. This renders:
\begin{eqnarray}\label{final-res}
-\sum_i a_i + \frac{1}{K} \sum_{i\ne j} a_i a_j - \frac{1}{K^2} \sum_{i\ne j \ne k} a_i a_j a_k \pm \dots  > \nonumber \\
-\sum_i b_i + \frac{1}{K} \sum_{i\ne j} b_i b_j - \frac{1}{K^2} \sum_{i\ne j \ne k} b_i b_j b_k \pm \dots \leftrightarrow \nonumber \\
\sum_i a_i + D_a < \sum_i b_i + D_b
\end{eqnarray}
using the correct individual signs for $D$. Now, for $K\rightarrow \infty$, the disturbance terms $D$ vanish on both sides and we get:
\begin{equation}\label{after_div}
\sum_i a_i < \sum_i b_i
\end{equation}
as needed.

\noindent
We conclude that for $K\rightarrow \infty$ the neural network is equivalent to the Bellman-Ford algorithm. 

In the following we will show by an extensive statistical evaluation that the distribution of $K$ for different realistic graph configurations is well-behaved and we never found any case where $K$ had to be larger than $10^6$.

Evidently $K$ \textit{can} grow if $\sum_i a_i \approx \sum_i b_i$. This can be seen easily as in this case we can express $\sum_i b_i = \sum_i a_i + \epsilon$, with $\epsilon$ a small positive number. When performing this setting we get from Eq.~\ref{final-res} that $\epsilon > D_b - D_a$, which can be fulfilled with large values for $K$.

\begin{figure}[ht!]
\centering \includegraphics[width=0.85\linewidth]{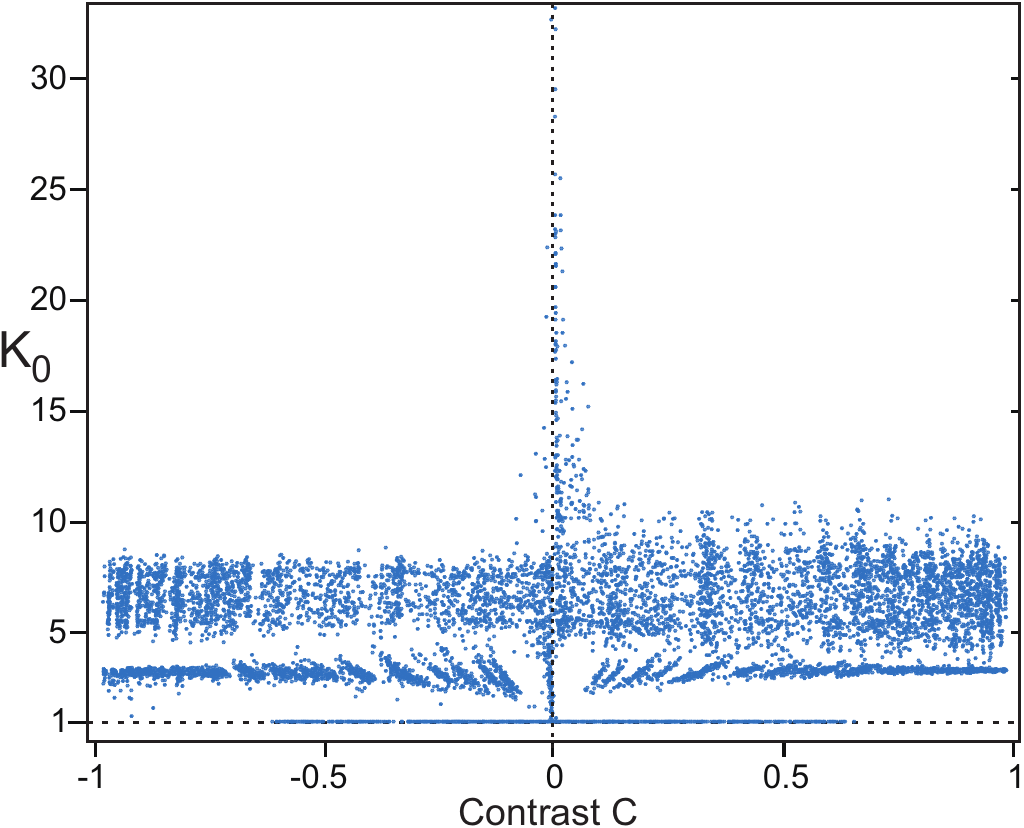}
\caption{Statistics for $K$ computed from a large number of 2-path combinations. Contrast $C=(\sum_i a_i - \sum_i b_i)/(\sum_i a_i + \sum_i b_i)$. For $K>K_0$, NN-BF renders identical results as BF.}
\label{K-stats}
\end{figure}

We have set the word "can" above in italics, because a broader analysis shows that -- even for similar sums -- $K$ gets somewhat larger only for very few cases only as can be seen from Fig.~\ref{K-stats}.

For this we performed the following experiment. We defined two paths A and B with different lengths $L_A$ and $L_B$, where $L$ was taken from $L\in \{2, 3, 4, 5, 6, 7, 8, 10, 20, 30, 40, 50, 100, 200 \}$ and all combinations of $L_A$ and $L_B$ were used. We evaluated 60 instances for each path pair (in total 11760 path pairs). To obtain those, we generated the cost for all edges in both paths using Gaussian distributions, where mean $m$ and variance $\sigma$ are chosen for each trial randomly and separately for A and B with $1\le m \le 21$ and $0\le \sigma \le 5$. 

In Fig.~\ref{K-stats}, we are plotting on the horizontal axis the "contrast" between the summed cost of path A and B given by $C={\frac{\sum_i a_i - \sum_i b_i}{\sum_i a_i + \sum_i b_i}}$. This allows comparing sums with quite different values. The vertical axis shows $K_0$, where for $K>K_0$ the neural network renders identical results as Bellman-Ford.

Note that we have truncated this plot vertically, but only $15$ values of $K_0$ had to be left out, all with contrast values $C \simeq 0$, where the largest was $K_0=812$ with a contrast of $C=5.3349\times 10^{-5}$. Hence, only for small contrast a few larger values of $K_0$ are found. Note also that the asymmetry with respect to positive versus negative contrast is expected, because fewer negative costs exist in paths A and B due to our choice of $1\le m \le 21$.

\begin{figure*}[ht!]
\centering \includegraphics[width=0.95\linewidth]{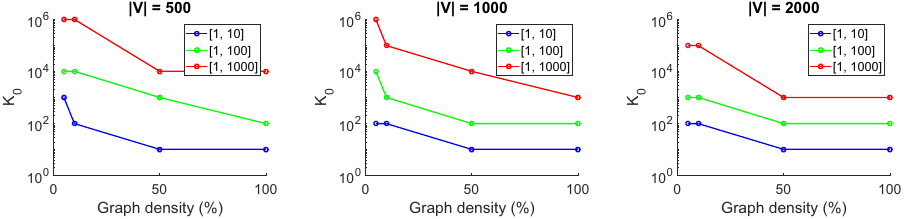}
\caption{Analysis of $K_0$ values for graphs of different sizes (500, 1,000, and 2,000), different densities (5, 10, 50 and 100\%) and different cost ranges ([1, 10], [1, 100], and [1, 1000]). For cases $|V| = 500$ with graph densities 5 and 10\% and cost range [1, 1000], and for $|V| = 1,000$ with graph density 5\% and cost range [1, 1000], 10,000 random graphs were used, whereas in all other cases 1,000 graphs were used.}
\label{K-pract}
\end{figure*}

\begin{figure}[ht!]
\centering \includegraphics[width=0.98\linewidth]{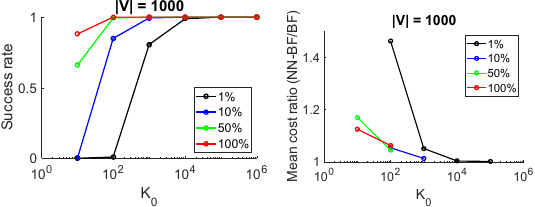}
\caption{Success rate of optimal solutions (left) and mean cost ratio (NN-BF cost / BF cost) of sub-optimal solution vs $K_0$ value. 1,000 randomly generated graphs with costs in the range [1,100] were used with four different densities (1, 10, 50, and 100\%). Note that for the graphs with 1\% density and with $K_0 = 10$ solutions were not found, and for the graphs with $10\%$ density and with $K_0 = 10$ solutions were found for four graphs and were optimal, whereas for all other remaining graphs solutions were not found.}
\label{K-conv}
\end{figure}

Considering only pairwise paths is of limited practical use. From such a perspective it is better to consider different graphs with certain cost ranges and different connection densities where one should now ask how likely it would be that results of BF and NN-BF are mismatched due to a possible $K$-divergence. To assess this we had calculated the statistics for three different graphs with sizes 500, 1,000, and 2,000 nodes and randomly created costs taken from three different uniform cost distributions with intervals: [1, 10], [1, 100], and [1, 1000]. We considered four different connectivity densities for each graph with 5, 10, 50, and 100\% connections each. Fig.~\ref{K-pract} shows the maximal $K_0$ found was $10^6$, hence that in all cases $K\ge 10^6$ will suffice, where such a high value is only needed for sparse graphs and large costs.

In addition, we have also performed experiments where we analysed the influence of the values of $K$ on the success rate for finding optimal solutions and on the cost ratio of the suboptimal solutions as compared to the solutions found by the BF algorithm. For this, we randomly generated 1,000 graphs with 1,000 nodes and four different densities: 1, 10, 50, and 100\% connections (fully connected graph), and cost range [1,1000]. Statistics are shown in Fig.~\ref{K-conv} where we show success rate of optimal solutions (left) and mean cost ratio (NN-BF cost / BF cost) of sub-optimal solutions for $K_0 = {10, 10^2, 10^3, 10^4, 10^5, 10^6}$. Results demonstrate that $K_0\geq 10^4$ values will lead to convergence to optimal solutions in all cases except for very sparse graphs (1 and 10\% connectivity), however, even in these cases the number of suboptimal solutions is very low (below 1\% and suboptimal solutions are very close to the optimal solutions (cost ratio $<1.004$). With $K_0=10^6$ optimal solutions were obtained in all cases. This shows that the choice of $K$ is non-critical in almost all cases.

\subsubsection{Algorithmic Complexity  of the Algorithms}

In the following we will show that all three algorithms have the same algorithmic complexity.

Bellman-Ford (BF) version 1 consists of two loops (see \cite{Baras2010}), where the outer loop runs in $\mathcal{O}(|V|)$ time and the inner loop runs in $\mathcal{O}(|E|)$, where $|V|$ and $|E|$ define the number of graph vertices (nodes) and edges (links), respectively. In the worst case, BF needs to be run for $|V|-1$ iterations and in the best case only for one iteration. Thus, the worst algorithmic complexity of BF is $\mathcal{O}(|V|\,|E|)$, whereas the best is $\mathcal{O}(|E|)$.

The algorithmic structure of the second version of the BF algorithm (see  \cite{Baras2010}) is the same as that of our NN-BF algorithm (Algorithm~\ref{alg_nn}). Hence their complexity is the same and we will provide analysis for the NN-BF algorithm. NN-BF operates on neurons and their inputs and not on edges of the graph as in BF version 1. Hence NN-BF is similar to BF version 2 that operates on the nodes. The outer loop, as in BF version 1, runs in $\mathcal{O}(N)$ where $N$ is the number of neurons and corresponds to the number of vertices $|V|$. The second loop iterates through all neurons and the third loop iterates through all inputs of the respective neuron and runs in $\mathcal{O}(N)$ and $\mathcal{O}(I_j)$, respectively. Here $I_j$ corresponds to the number of inputs of a particular neuron $j$. Given the fact that $N = |V|$ and $\sum_j{I_j} = |E|$ ($j = 1 \dots N$), we can show that the worst and the best algorithmic complexity of the NN-BF algorithm is the same as for BF -- version 2, i.e., $\mathcal{O}(|V|\,|E|)$ and $\mathcal{O}(|E|)$, respectively. Hence, this is the same complexity as for BF-version 1, too.

\subsubsection{Convergence analysis}

\begin{figure}[ht!]
\centering \includegraphics[width=0.6\linewidth]{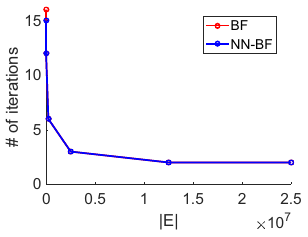}
\caption{Results of convergence analysis. Maximal number of iterations until convergence of Bellman-Ford -- version 1 (BF) or NN-BF  vs. number of edges $|E|$. 500 randomly generated graphs with 5,000 nodes were analysed for each case.}
\label{res_iter}
\end{figure}

The worst case scenario would be to run the algorithms for $|V|-1$ iterations where $|V|$ is the number of graph nodes. However, in practice this will be not needed and significantly fewer iterations will usually suffice. Thus, we tested how many iterations are needed until cost convergence of the Bellman-Ford algorithm (version 1) as compared to activation convergence of the neural network. In both cases we stopped the respective algorithm, as soon as its node costs or neuron activations were not changing anymore.

For this analysis, we used 500 randomly generated graphs with 5,000 nodes with positive costs ($p_n = 0$) from the range $[1,10]$ and connectivity densities corresponding to approximately $12.5\times 10^3$, $25\times 10^3$, $250\times 10^3$, $2.5\times 10^6$, $12.5\times 10^6$, and to $24.995\times 10^6$ edges.

The results are shown in Fig.~\ref{res_iter} where we show the maximal number of iterations obtained from 500 tested graphs until convergence. One can see that in case of sparse graphs 16 for BF and 15 for NN-BF iterations suffice for convergence and for denser graphs ($>10\%$ connectivity) only 3 iterations are needed. The resulting run-time per one iteration on a sparse graph with 5,000 nodes and 12,500 edges is 21$\mu s$ and 59$\mu s$ for BF (version 1) and NN-BF, respectively. Run-time on a fully connected graph with 24,995,000 edges is 26$ms$ and 14$ms$ for BF and NN-BF, respectively\footnote{C++ CPU implementation on Intel Core i9-9900 CPU, 3.10GHz, 32.0GB RAM.}. Thus, relatively large graphs can be processed in less than $100~ms$.

\subsection{Network learning rule}

As described above, the shortest path in the network between a source and a target neuron is given by that specific sequence of connected neurons, which leads to the largest activation at the target neuron. Hence, learning that modifies connection weights (and, thus, also the resulting activations) can alter path-planning sequences.

In this section we, thus, define a Hebbian type synaptic learning rule to modify connection weights \cite{Hebb1949,Porr2003,Porr2007}, which will later be used in some examples.

In general, Hebbian plasticity is defined as a multiplicative process where a connection between two neurons --- its synapse --- is modified by the product of incoming (pre-synaptic) with outgoing (postsynaptic) activity of this neuron \cite{Hebb1949}. Often a third factor $R$, by means of an additional incoming activation enters this correlation process also in a multiplicative way to provide additional control over the weight change leading to a three-factor rule (see \cite{Kusmierz2017} for a discussion of three-factor learning). We now first present our three-factor Hebbian learning scheme and rule, and then we present two learning scenarios, namely, navigation learning and sequence learning.  

\begin{figure}[ht!]
\centering \includegraphics[width=0.75\linewidth]{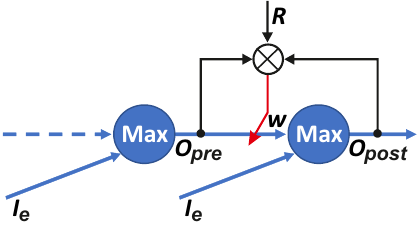}
\caption{Schematic diagram of the learning architecture. $I_e$ - external input; $R$ - reward signal; $O_{pre}$ and $O_{post}$ - output of pre-synaptic and post-synaptic neuron, respectively; $w$ - synaptic weight between pre- and post-synaptic neuron. The multiplication symbol followed by a red arrow indicates that the synaptic weight $w$ is changed by a three-factor multiplicative (correlation) process. (see Eq.~\ref{learningrule}).}
\label{learn_diag}
\end{figure}

To allow for learning (see schematic diagram in Fig.~\ref{learn_diag}), neurons will also receive additional external input $I_e = 1$ signaling occurrence of specific events, for instance, if an agent is at the specific location in an environment (otherwise it will be set to $I_e = 0$). The network will receive external inputs $I_e$ only during learning but not during planning.
The output of a  post-synaptic neuron, $O_{post}$, is computed as described above using Eq.(\ref{eq_out}). Hence, $O_{post} = max\{O_{pre}, I_e\}$.

The synaptic weight $w$ between pre- and post-synaptic neuron is then changed according to
\begin{equation}
 \Delta w = \alpha \, O_{pre} \, O_{post} \, R,
\label{learningrule}
\end{equation}
where $\alpha>0$ is the learning rate, usually set to small values to prevent too fast learning. The driving force of the learning process is the reward signal $R$ which can be positive or negative and will lead to weight increase (long-term potentiation [LTP]) or decrease (long-term depression [LTD]), respectively. This learning rule represents, thus, a three-factor rule \cite{Kusmierz2017}, where here the third-factor is the reward signal $R$.

A diagram of the learning scheme is presented in Fig.~\ref{learn_diag}, where we show components of the learning rule to change the synaptic weight between two neurons in the planning network.

Note that in our learning scheme weights were bounded between $0$ and $1$, i.e., $0 \leq w < 1$.

\subsection{Combining Learning with Path Planning - Two Examples}

In the following we will show examples of network learning combined with path finding. Examples stem from two common problem sets (navigation learning and sequence learning) and we can demonstrate here the main advantage of BF-NN, which is that learning and path finding relies on the same numerical representation. To achieve the same with BF (or any other additively operating algorithm) one would have to transform costs (for path finding) to activations (for network learning) back and forth. Examples are kept small, but upscaling to large problems is straight forward.

\subsubsection{Navigation learning}

In the first learning scenario, we considered a navigation task in 2D space, where an agent had to explore an unknown environment and plan the shortest path from a source location to a destination.

\begin{figure*}[ht!]
\centering \includegraphics[width=0.95\linewidth]{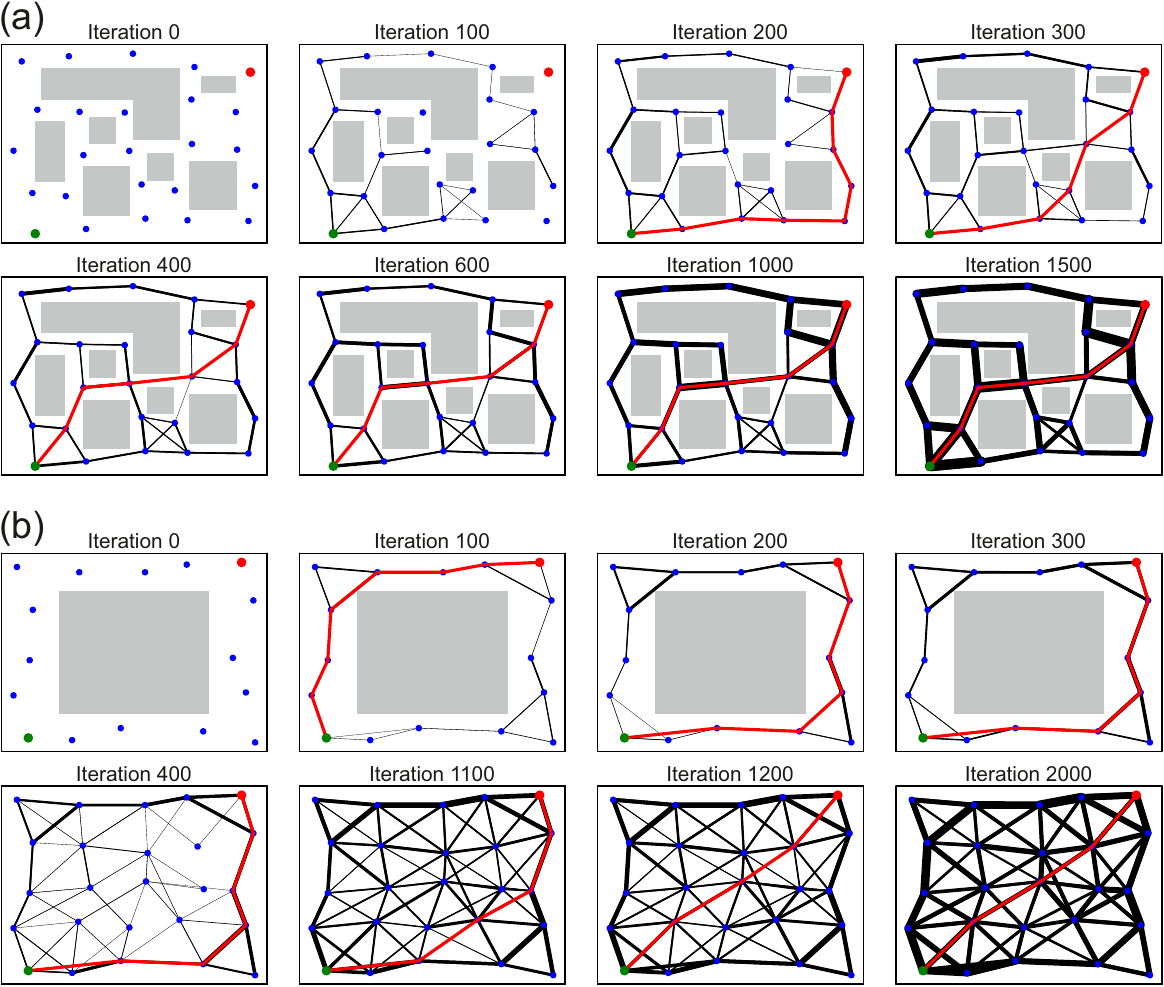}
\caption{Results of navigation learning: (a) static environment and (b) dynamic environment. Blue dots correspond to locations represented by neurons, and black lines correspond to connections between neurons (paths between locations) where line thickness is proportional to the synaptic weight strength. Grey blocks denote obstacles. Green and red dots correspond to the start- and the end-point of the path marked by the red trajectory.}
\label{res_nav}
\end{figure*}

We assume a rectangular arena, see Fig.~\ref{res_nav}(a,b), iteration 0, with some locations (blue dots) and obstacles (grey boxes). The locations are represented by neurons in the planning network, whereas connections between neurons correspond to possible transitions from one location to other neighbouring locations. Note that some transitions between neighboring locations, and, thus, connections between neurons are not possible due to obstacles. The positions of locations and possible transitions between them were generated in the following way. The position of a location $k$ is defined by
\begin{equation}
(x_k,y_k) = (h\,i+\xi_i, h\,j+\xi_j),
\end{equation}
where $i,j = 1, 2, \dots m$, and $\xi$ is noise from a uniform distribution $\mathcal{U}(a,b)$. In this study, we used $m=5$ ($k=m^2=25$), $h=0.2$, $a=-0.05$, and $b=0.05$.

Possible transitions between neighbouring locations, and, thus, possible connections between neurons were defined based on a distance threshold $\theta_d$ between locations, i.e., connections were only possible if the Euclidean distance between two locations $d_{k,l}=||(x_k,y_k)-(x_l,y_l)|| \leq \theta_d$. In one case (Fig.~\ref{res_nav}(a)) we used $\theta_d=0.25$ and in another case (Fig.~\ref{res_nav}(b)) we used $\theta_d=0.35$.

Synaptic weights between neurons were updated after each transition (called iteration) from the current location $k$ (pre-synaptic neuron) to the next neighboring location $l$ (post-synaptic neuron) according to the learning rule as described above, where pre- and post-synaptic neurons obtain an external input $I_e=1$ during this transition (otherwise $I_e=0$). In the navigation scenario the reward signal $R$ was inversely proportional to the distance between neighbouring locations $d_{k,l}$, i.e., $R = 1 - \beta \, d_{k,l}$ (here we used $\beta=1$). Thus, smaller distances between locations lead to larger reward values and larger weight updates, and vice versa. In the navigation learning scenario, initially all weights were set to $0$, and the learning rate was set to $\alpha = 0.02$. The learning and planning processes were repeated many times, where the learning process was performed for 100 iterations and then a path from the source (bottom-left corner) to the destination location (top-right corner) was planned using current state of the network.

We used two navigation learning cases. In the first case, a static environment with multiple obstacles was used where obstacles were present during the whole experiment (see Fig.~\ref{res_nav}(a)), whereas in the second case a dynamic environment was simulated, where one big obstacle was used for some period of time and then was removed some time later (see Fig.~\ref{res_nav}(b)).

Results for  navigation learning in a static environment are shown in Fig.~\ref{res_nav}(a), where we show the development of the network's connectivity (black lines; line thickness is proportional to the weight strength) and the planned path (red trajectory) based on the network's connectivity during the learning process. After 100 learning iterations, the environment was only partially explored. Therefore, the full path could not yet be planned. After 200 iterations, only a sub-optimal path was found, since the environment was still not completely explored. As exploration and learning process proceeds, connections between different locations get stronger (note thicker lines in the second row), and eventually the planned path converges to the shortest path (after 400 iterations).

In Fig.~\ref{res_nav}(b), we show results for  navigation in a dynamic environment. Here, we first run the learning process until the path has converged (see the top row), and then remove the obstacle after 300 learning iterations. As expected, after obstacle removal, the middle part of the environment is explored and new connections between neurons are built. Eventually, as the learning proceeds, the output of the planning network converges to the shortest path (after 1200 iterations).

Due to the learning rule used, in both cases, systems converge to the shortest euclidean distance paths.

\begin{figure*}[ht!]
\centering \includegraphics[width=0.95\linewidth]{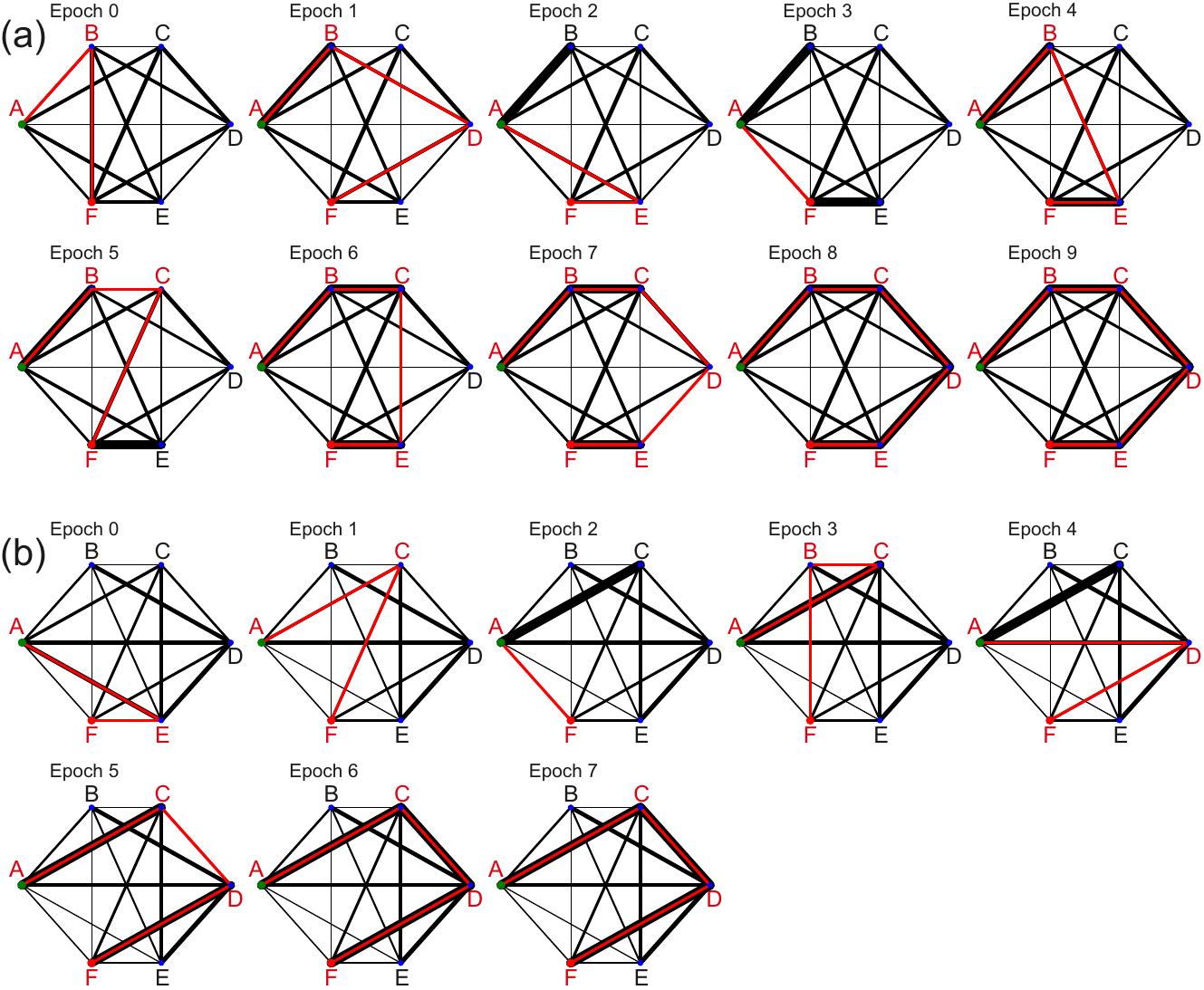}
\caption{Results of sequence learning: (a) learning of sequence $A\,B\,C\,D\,E\,F$, and (b) learning of sequence $A\,C\,D\,F$. Blue dots correspond to letters represented by neurons, and black lines correspond to connections between neurons where line thickness is proportional to the synaptic weight strength. Green and red dots correspond to the start- and end-point of the sequence marked by the red trajectory.}
\label{res_seq}
\end{figure*}

\subsubsection{Sequence learning}

In the second learning scenario, we were dealing with sequence learning, where the task was to learn executing a predefined sequence of arbitrary events, which in our study were denoted by letters. For instance, this could correspond to an action sequence in order to execute some task by a robot, and could be used for task planning in robotics to generate optimal plans. Note, that sequences do not necessary have to be predefined (as in our example), but optimal sequences can be learnt based on obtained rewards during action execution.

In this case, we used a fully connected network (without self-connections) were neurons correspond to letters. In total we used six different events (letters), i.e., $A$, $B$, $\dots$, $F$ (see Fig.~\ref{res_seq}(a)). Initially all weights were set to random values from a uniform distribution $\mathcal{U}(0,0.5)$. Here, the planning and learning procedure is performed in epochs, where we execute the sequence based on the planner's outcome always starting with letter $A$ and ending with letter $F$, and perform weight updates for each pair in the planned sequence. Suppose that at the beginning the planner generates a sequence $A\,B\,F$, then there will be two learning iterations in this epoch, i.e., weight update for the transition from $A$ to $B$ and from $B$ to $F$. As in the navigation learning scenario, pre- and post-synaptic neurons will receive external input $I_e=1$ whenever a certain transition from one event to the next event happens, e.g., from $A$ (pre-synaptic neuron) to $B$ (post-synaptic neuron).

Different from the navigation scenario, here we used positive and negative rewards, i.e., $R=1$ if a sequence-pair is correct and $R=-1$ otherwise. In case of our previous $A\,B\,F$ example, this would lead to an increase of the synaptic weight between neurons $A$ and $B$ (correct), and to a decease of the synaptic weight between neurons $B$ and $F$ (incorrect; after $B$ should be $C$). The learning procedure is repeated for several learning epochs until convergence. Here we used the relatively high learning rate $\alpha=0.9$, which allows fast convergence to the correct sequence.

We performed two sequence learning cases where in the first case we were learning a sequence  consisting of all possible six events $A\,B\,C\,D\,E\,F$, and in the second case a shorter sequence $A\,C\,D\,F$ had to be learnt.

Results for learning of the full sequence $A\,B\,C\,D\,E\,F$ and of the partial sequence $A\,C\,D\,F$ are presented in Fig.~\ref{res_seq}(a) and Fig.~\ref{res_seq}(b), respectively. As in the navigation learning example, we show the development of the network's connectivity during the learning process. In case (a), we can see that before learning (epoch 0), due to random weight initialisation, the generated sequence is $A\,B\,F$ which is neither correct nor complete. However, after the first learning epoch we can see that the connection weight between neurons $A$ and $B$ was increased (note thicker line between $A$ and $B$ as compared to epoch 0), whereas the connection weight between neurons $B$ and $F$ was decreased, which then led to a different sequence $A\,B\,D\,F$. After epoch 3, the connection weight between neurons $E$ and $F$ was increased since in previously generated sequence $A\,E\,F$ (epoch 2) the transition from $E$ to $F$ was correct. Finally, after learning epoch 7 the network generates the correct sequence.

Similarly, in case of learning of the partial sequence $A\,C\,D\,F$ (see Fig.~\ref{res_seq}(b)),  learning converges to the correct sequence already after epoch 5, since the sequence is shorter.   

\section{Conclusion and outlook}
Finding the optimal path in a connected graph is a classical problem and, as discussed in the section on state of the art, many different algorithms exist to address this. Here we proposed a neural implementation of the Bellman-Ford algorithm. The main reason, as we would think, for translating Bellman-Ford into a network algorithm is the now-arising possibility to directly use network learning algorithm on a path finding problem. To do this with BF or other cost-based algorithm, you would have to switch back and forth between a cost- and a weight-based representation of the problem. This is not needed for NN-BF. Given that BF and NN-BF have the same algorithmic complexity, computational speed of the path-finding step is similar, too, where learning epochs can be built into NN-BF as needed. The examples shown above rely on a local learning rule that alters the weights only between directly connected nodes. Globally acting rules could, however, also be used to address different problems, too, and -- as far as we see -- there are no restrictions with respect to possible network learning mechanisms, because BF-NN is a straight-forward architecture for activity propagation without any special limitations.

Hence, network algorithm with NN-BF allows for new and different applications where learned path-augmentation is directly coupled with path finding in a natural way.

So far we only investigated the performance of our method on specific environments and tasks. Some of these task addressed the problem of unseen environments using our learning methods. In our future work we will investigate the capability of our approach to generalize across different environments and/or different tasks and we can also extend the experiments into a deeper investigation of complex unknown environments to even more strongly emphasise the power of being able to combine path finding directly with learning.




\ifCLASSOPTIONcaptionsoff
  \newpage
\fi


\bibliographystyle{IEEEtran}
\bibliography{Kulvicius_et_al_TNNLS_2022_arxiv_2.bib}

\end{document}